\documentclass[letterpaper, 10pt, journal, twoside]{IEEEtran}
 \usepackage[utf8]{inputenc}

\usepackage[exponent-product=.,detect-weight]{siunitx}
\usepackage[hyphens]{url}
\usepackage[breaklinks]{hyperref}
\hypersetup{hidelinks}
\usepackage{amsmath, 
	amsfonts, 
	amssymb, 
	bm,
	mathtools,
	amsthm,
	siunitx}
\usepackage{graphicx}
\usepackage[dvipsnames]{
		xcolor
	}
\usepackage{booktabs}
\usepackage{multirow}
\usepackage{tikz,
	pgfplots,
	pgfplotstable,
	eso-pic}
\usetikzlibrary{patterns,
	plotmarks}
\pgfplotsset{compat=1.12}

\usepackage[english,
	ruled,
	]{algorithm2e}
	
\usepackage[nospace]{cite}

\usepackage{blindtext,
	ifpdf}
\usepackage{setspace}
\usepackage{paralist}
\usepackage{filecontents}
\usepackage{ifthen}

\newcommand{\bfA}{\mathbf{A}}
\newcommand{\bfb}{\mathbf{b}}

\newcommand{\bfB}{\mathbf{B}}

\newcommand{\bfG}{\mathbf{G}}

\newcommand{\bfH}{\mathbf{H}}
\newcommand{\bfI}{\mathbf{I}}
\newcommand{\bfJ}{\mathbf{J}}

\newcommand{\calN}{\mathcal{N}}

\newcommand{\bfP}{\mathbf{P}}

\newcommand{\calP}{\mathcal{P}}

\newcommand{\bfQ}{\mathbf{Q}}
\newcommand{\hbfQ}{\mathbf{\hat{Q}}}
\newcommand{\calQ}{\mathcal{Q}}

\newcommand{\bfR}{\mathbf{R}}

\newcommand{\bbR}{\mathbb{R}}

\newcommand{\bft}{\mathbf{t}}

\newcommand{\bfT}{\mathbf{T}}
\newcommand{\hbfT}{\mathbf{\hat{T}}}

\newcommand{\bfw}{\mathbf{w}}

\newcommand{\bophi}{\boldsymbol{\phi}}

\newcommand{\borho}{\boldsymbol{\rho}}

\newcommand{\boxi}{\boldsymbol{\xi}}

\newcommand{\hboxi}{\mathbf{\hat{\boxi}}}

\newcommand{\bfzero}{\mathbf{0}}

\newcommand{\icp}{\mathrm{icp}}
\newcommand{\true}{\mathrm{true}}

\newcommand{\ini}{\mathrm{ini}}
\newcommand{\rel}{\mathrm{rel}}

\DeclareMathOperator{\cov}{cov}
\DeclareMathOperator{\rmicp}{icp}
\DeclareMathOperator{\col}{col}

\DeclareMathOperator{\trace}{trace}

\newtheoremstyle{mystyle}
{}
{}
{\itshape}
{}
{\bfseries}
{.}
{ }
{}

\usetikzlibrary{shapes,arrows,shadows,snakes,automata,backgrounds,petri}
\pgfdeclarelayer{background}
\pgfdeclarelayer{foreground}
\pgfsetlayers{background,main,foreground}

\tikzstyle{wa} = [draw, text width=10em, fill=red!20,
minimum height=4em, rounded corners, drop shadow, text centered]
\tikzstyle{de} = [draw, text width=10em, fill=blue!20,
minimum height=4em, rounded corners, text centered,drop shadow]

\usepackage{afterpage}
\usepackage{amsmath,bm,times}
\usepackage{filecontents}
\usepackage{pgfplots,pgfplotstable}
\usepackage[top=54pt, bottom=54pt, left=54pt, right=54pt]{geometry}

\pgfplotsset{grid style={dashed,gray}}
\pgfplotsset{major grid style={dashed,gray}}
\pgfplotsset{minor grid style={dotted,gray!50}}

\author{Martin Brossard$^{1}$, Silvère Bonnabel$^{1}$, and Axel Barrau$^{2}$%
\thanks{Manuscript received: September, 10, 2019; Revised November, 1, 2019;
Accepted December, 23, 2019.}
\thanks{This paper was recommended for publication by Editor 
Sven Behnke upon
evaluation of the Associate Editor and Reviewers' comments.}
\thanks{$^{1}$Martin Brossard and Silvère Bonnabel are with MINES ParisTech, PSL Research University, Centre for Robotics, 60 Boulevard Saint-Michel, 75006 Paris, France
{\tt\small \{martin.brossard, silvere.bonnabel\}@mines-paristech.fr}}%
\thanks{$^{2} $Axel Barrau is with Safran Tech, Groupe Safran, Rue des Jeunes Bois-Ch\^ateaufort, 78772, Magny Les Hameaux Cedex, France
{\tt\small axel.barrau@safrangroup.com}}%
\thanks{Digital Object Identifier (DOI): see top of this page.}
}

\begin{document}
\bstctlcite{IEEEexample:BSTcontrol}
\title{\vspace*{0.0cm} A New Approach to 3D ICP Covariance Estimation  }

\maketitle

\begin{abstract}
In mobile robotics, scan matching of point clouds using Iterative Closest Point
(ICP) allows estimating   sensor displacements. It may prove important to assess
the associated uncertainty about the obtained rigid transformation, especially
for sensor fusion purposes. In this paper we propose a novel approach to 3D uncertainty of ICP  that accounts for all the sources of error as listed in
Censi's pioneering work \cite{censiAccurate2007}, namely wrong convergence, underconstrained situations,
and sensor noise. Our approach builds on two facts. First, the uncertainty about the ICP's output fully depends on the
initialization accuracy. Thus speaking of the covariance of ICP makes   sense only in relation to the  initialization uncertainty, which generally stems from odometry errors.  We capture this using the unscented transform, which also reflects
correlations between initial and final uncertainties. Then, assuming white
sensor noise leads to overoptimism as ICP is biased owing to e.g. calibration
biases, which we account for. Our solution is tested on  publicly available real
data ranging from structured to unstructured environments, where our algorithm
predicts consistent results with actual uncertainty, and compares   favorably
to previous methods.
\end{abstract}


\section{Introduction}\label{sec:int}

\IEEEPARstart{P}{oint} clouds and the Iterative Closest Point (ICP) algorithm play a crucial role
for localization and mapping in modern mobile robotics
\cite{pomerleauReview2015, holzRegistration2015}. ICP computes an estimate of
the 3D rigid transformation that aligns a reading point cloud to a reference
point cloud (or more generally a  model  or a surface). The algorithm starts
with a first transformation estimate, and repeats - until convergence - point
association and least-square minimization, where 
initialization is naturally provided in mobile robotics by odometry
\cite{dubeOnline2017,genevaLIPS2018} based on wheel speeds,  inertial sensors,
or vision. The point association matches points between the two clouds by
generally associating each point of the second cloud to its closest point in the
first one. Then,
the algorithm minimizes a user-chosen metric between the matched
points that provides an update of the current estimate. In
spite of  robust filtering that
are broadly used during the alignment of point clouds, a.k.a. registration, ICP
is subject to errors stemming from sensor noises, underconstrained environments
that result in unobservable directions, and  local minima \cite{pomerleauNoise2012,censiAccurate2007,landryCELLO3D2019}.

\subsection{Sources of ICP Uncertainty }\label{source:sub}

\begin{figure}
\centering
\includegraphics{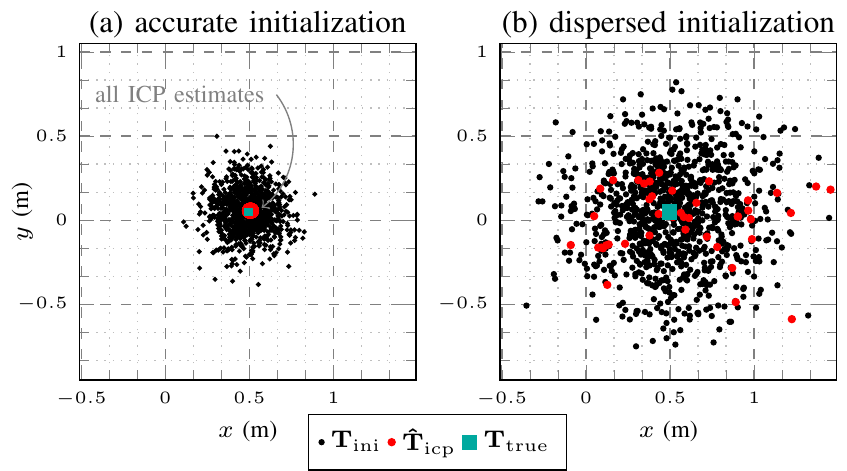}
\vspace*{-0.5cm}
\caption{\label{fig:Q_init}Horizontal translation estimates according to ICP
($\hbfT_{\rmicp}$, red dots)
for various  initial estimates ($\bfT_{\ini}$, black dots) and
ground-truth ($\bfT_\true$, square) for registering two scans of the sequence
  \emph{Stairs} of \cite{pomerleauChallenging2012},
  where we sample \num{1000} initial estimates from two distributions
  reflecting accurate (a) and dispersed (b) ICP initialization and that
  respectively correspond to the easy and medium scenarios of
  \cite{pomerleauComparing2013}. We see the uncertainty on the ICP estimate,
  that is,   dispersion of red points, wholly depends on the accuracy of
  initialization. There is no   ``uncertainty of ICP''  \emph{per se}.\vspace*{-0.5cm}}
\end{figure}

The pioneering work of Censi  \cite{censiAccurate2007}  identifies the following
sources of error for ICP registration: wrong convergence (not handled by Censi's
formula), underconstrained situations, and sensor noise. As indicated by
preliminary remarks in  \cite{pfisterWeighted2002,
barczykRealistic2017} we believe a fourth important source
is missing: the one that stems from sensor biases. In the present paper we
consider indeed  the following sources of error:
\subsubsection{Initial Transformation} ICP is subject to error due to wrong initialization that makes the algorithm converge to a local minimum out of the attraction basin of the true solution, as largely observed in practice, see e.g. \cite{landryCELLO3D2019,iversenPrediction2017} and Figure \ref{fig:Q_init}. In practice it often proves to be the dominant error.
\subsubsection{Sensor White Noise} each \emph{point} measured in a point cloud is affected by an \emph{independent} random sensor noise of  centimetric magnitude which is  a  function of point depth and beam angle\cite{pomerleauNoise2012, wangCharacterization2018}.
\subsubsection{Sensor Bias Noise} the observed points share common errors that stem from: temperature drift effect, i.e. stability of the laser \cite{wangCharacterization2018}; observed material \cite{pomerleauNoise2012}; incidence and beam angles resulting in large bias   \cite{laconteLidar2019}; or wrong calibration, e.g. \cite{deschaudIMLSSLAM2018} found a distortion of \SI{0.22}{\deg} of the scan point clouds due to  intrinsic calibration process. This \emph{correlated} noise, a.k.a. bias, strictly limits the confidence we may have in the ICP estimate. To our best knowledge this is often omitted with a few exceptions: e.g., \cite{laconteLidar2019} removes   bias on point measurements due to sensor beam angle,  and preliminary ideas may be found in  \cite{pfisterWeighted2002,barczykRealistic2017}.

\subsubsection{Randomness Inherent to the ICP Algorithm} ICP is generally
configured with random filtering processes \cite{pomerleauReview2015}, e.g. sub-sampling,
such that two solutions with exactly the same inputs would differ.

In the following we address uncertainty coming from  1), 2) and 3) and do not
consider 4), which should be marginal.

\subsection{Brief Literature Review}
Various approaches exist for estimating the covariance of the ICP algorithm,
each of which being a trade-off between accuracy and execution time. Monte-Carlo
algorithms, e.g. \cite{iversenPrediction2017,bengtssonRobot2003}, sample  noisy
scans (from a reference scan) and ICP initializations to compute a large number
of ICP registration results, define the covariance of the sampled results as the
covariance estimation, and use the estimated covariance for all future
registration with the reference scan, thus getting a covariance function of the
reference scan only. Another category of covariance estimation methods relies on
closed-form expressions \cite{biberNormal2003,
censiAccurate2007,bonnabelCovariance2016, prakhyaclosedform2015}, whose underlying assumption consists in linearizing the
objective function used in ICP around the convergence point, ruling out the
possibility for wrong convergence and the uncertainty that stems from it.
Albeit still used in practice, Censi's pioneering formula
\cite{censiAccurate2007} is widely considered as overoptimistic, see e.g.
\cite{mendesICPbased2016}. Recently,
\cite{landryCELLO3D2019} leveraged learning based approaches to estimate ICP
uncertainty stemming from inaccurate ICP initialization.

\subsection{Contributions and Paper's Organization}

Our approach introduced in  Section \ref{sec:source} extends  existing works in three ways: 1) we consider ICP uncertainty coming both from sensor errors and ICP initialization. 2) we raise an important point  which is that ICP uncertainty in itself is meaningless as it is inherently related to uncertainty in the initialization pose  (unless there is a global minimum). This is supported by experiments displayed in Figure \ref{fig:Q_init}. We address this problem by outputting a covariance matrix of larger dimension that also reflects the correlation between ICP final and initial estimates. And 3)  we estimate in Section \ref{sec:cova} the ICP uncertainty combining a closed-form expression using \cite{censiAccurate2007,bonnabelCovariance2016} accounting for sensor biases,   and derivative-free methods using the unscented transform of  \cite{juliernew2000,brossardUnscented2017}, which comes at a lower computational cost than Monte-Carlo runs.

Besides, we evaluate, compare and discuss our approach on the dataset of
\cite{pomerleauChallenging2012} in Sections \ref{sec:results} and \ref{sec:results2}, where our
approach obtains consistent estimates and achieves better results than existing
methods.
The code to reproduce the results of the paper is made publicly available at: \texttt{\url{https://github.com/CAOR-MINES-ParisTech/3d-icp-cov}}.

Throughout the article, we configurate the ICP as suggested in \cite{pomerleauComparing2013} with a point-to-plane error metric.

\section{Proposed Approach}\label{sec:source}

\subsection{Pose Representation and Pose Uncertainty Representation}\label{sec:pos}

The true transformation between  two point clouds and its ICP-based estimate both live in the set of 3D rigid transformations
\begin{align*}
SE(3):= \left\{ \bfT= \begin{bmatrix}
\bfR & \bft \\ \bfzero & 1
\end{bmatrix} \in \bbR^{4\times4}|\bfR\in SO(3), \bft \in \bbR^3 \right\},
\end{align*}
and are thus represented by a matrix $\bfT$ (a.k.a. homogeneous coordinates), where $\bfR$ denotes a rotation matrix and $\bft$ a translation. Note that, it is consistent with matrix multiplication: if $\bfT_1$ transforms a first point cloud into a second one, and then $\bfT_2$ transforms the latter into a third cloud, then the matrix $\bfT_2 \bfT_1 \in SE(3)$ encodes the transformation between the first and the third  clouds.

It is possible to linearize poses through the approximations $\cos(\alpha) \simeq 1$ and $\sin(\alpha)\simeq \alpha$ for small $\alpha$. For example, we have for a small rotation around the $x$ axis of angle $\alpha$
\begin{align}
\bfR_{x,\alpha}&=\begin{bmatrix}
1 & 0&0 \\ 0 & \cos(\alpha) & -\sin(\alpha) \\ 0 & \sin(\alpha) & \cos(\alpha)
\end{bmatrix} \simeq  \begin{bmatrix}
1 & 0& 0 \\ 0 &1 &-\alpha \\ 0 & \alpha & 1
\end{bmatrix}  \nonumber\\
&\simeq \bfI_3 + \begin{bmatrix}
0 & 0& 0 \\ 0 &0 &-\alpha \\ 0 & \alpha & 0
\end{bmatrix} = \bfI_3 + \alpha (\begin{bmatrix}
1\\0\\0
\end{bmatrix})_\times,
\end{align}
where $(\bfb)_\times \in \bbR^{3\times3}$ denotes the skew symmetric matrix
associated with cross product with $\bfb \in \bbR^3$. Along those lines, a full
rotation $\bfR$ may be approximated as
\begin{align}
\bfR \simeq \begin{bmatrix}
1 &-\gamma & \beta \\ \gamma & 1& -\alpha \\ -\beta & \alpha &1
\end{bmatrix} =\bfI_3 + (\begin{bmatrix}\alpha \\ \beta \\ \gamma \end{bmatrix})_\times
\end{align}
for small rotations around the $x$, $y$ and $z$ axes. The identity pose writes $Id = \bfI_4$ and a transformation being close to identity may thus be linearized as $\bfT \simeq \bfI_4 + \boxi^\wedge$ with
\begin{align}
\boxi^\wedge: = \begin{bmatrix}
(\bophi)_\times & \borho \\
\bfzero & 0
\end{bmatrix} \in \bbR^{4\times 4}, \boxi = \begin{bmatrix}\bophi\\ \borho\end{bmatrix}, \bophi \in \bbR^3, \borho \in \bbR^3.
\end{align}
This may serve as an uncertainty representation for poses as follows. If $\boxi$
is taken random, typically we take a Gaussian $\boxi \sim
\calN\left(\bfzero,\bfQ\right)$, where $\bfQ\in\bbR^{6\times6}$ is the
covariance matrix, then $\bfI_4+\boxi^\wedge$ defines a small random pose close
to identity, i.e., a small transformation. In turn,  for a given
pose $\bfT$, the transformation $\hbfT = \bfT\left(\bfI_4+\boxi^\wedge\right)
= \bfT + \bfT\boxi^\wedge$ denotes a random transformation being close to
$\bfT$. $\bfT$ may be viewed as the noise free mean of the random pose $\hbfT$, and
$\bfQ$ encodes the dispersion around the mean value $\bfT$.

A further theoretical step in this direction consists in using the notion of
concentrated Gaussian distribution as advocated in \cite{barfootAssociating2014}, see also
\cite{barrauInvariant2018,barrauInvariant2017,bourmaudContinuousDiscrete2015},
\begin{align}
	\hbfT = \bfT\exp(\boxi)\text{, where } \boxi \sim \calN\left(\bfzero,\bfQ\right), \boxi\in \bbR^6, \label{eq:T}
\end{align}
with $\boxi$ a zero-mean Gaussian variable of covariance $\bfQ$ and where $\exp(\cdot)$ denotes the exponential map of $SE(3)$. The latter maps elements $\boxi$ to poses. Albeit sounder from a
mathematical standpoint, this is very close to what we have just presented
since the $\exp(\cdot)$ map has the property that  $\exp(\boxi) \simeq \bfI_4 +
\boxi^\wedge$ up to first order term in $\boxi$. For
uncertainty representation \eqref{eq:T}, we adopt the notation $\hbfT \sim
\calN_L(\bfT,\bfQ)$.

Note that, in \eqref{eq:T}, the vector $\boxi\in\mathbb R^6$ may be viewed as the
\emph{error} between $\bfT$ and $\hbfT$. Indeed the relative transformation between poses $\bfT$ and $\hbfT$ is encoded in
$\boxi$ as
$\bfT^{-1}\hbfT=\exp(\boxi)$.

\subsection{The Role of ICP Initialization}\label{sec:rol}

The ICP procedure seeks to estimate the transformation $\bfT_{\true} \in
SE(3)$ that maps a first cloud of points $\calP$ to a second cloud (or a model) $\calQ$ as
follows \cite{pomerleauReview2015, holzRegistration2015}:
\begin{enumerate}
\item[$i$)] we have a first ``guess" for the transformation we call $\bfT_{\ini}$,
a.k.a. initial or coarse alignment \cite{holzRegistration2015};
\item[$ii$)] then we initialize the ICP algorithm by applying a
transformation $\bfT_{\ini}^{-1}$ to the cloud $\calQ$. This way the
transformation the ICP seeks to estimate become the relative pose $\bfT_{\rel}: =
\bfT_{\ini}^{-1}\bfT_{\true}$. We thus get   an estimate
$\hbfT_{\rel} = \icp\left(\calP,  \bfT_{\ini}  ^{-1}
\calQ\right)$ for $\bfT_{\rel}$;
\item[$iii$)] finally the estimate of   $\bfT_{\true}$ that
the algorithm outputs is
$\hbfT_{\icp}:=\bfT_{\ini} \hbfT_{\rel}=\bfT_{\ini} \icp\left(\calP,  \bfT_{\ini}  ^{-1}
\calQ\right)$.
\end{enumerate}
Note that \emph{if} $\bfT_{\rel}$ is
perfectly estimated we recover $\bfT_{\true}$ as then $\hbfT_{\icp}=\bfT_{\ini} \bfT_{\rel}=\bfT_{\ini}\bfT_{\ini}^{-1}\bfT_{\true}=\bfT_{\true}$ no matter how far the initial guess $\bfT_{\ini}$ is from $\bfT_{\true}$.

Let us introduce the various errors at play in ICP. In robotics, the initial
guess in $i$) is typically provided
through inertial sensors or wheeled odometry \cite{dubeOnline2017,genevaLIPS2018}. We have thus an
initialization error  that stems from sensor imperfections, encoded by a vector
$\boxi_{\ini}$, and one may
write
\begin{align}
	\bfT_{\ini} = \bfT_{\true}\exp(\boxi_{\ini}), ~ \boxi_{\ini} \sim \calN(\bfzero,\bfQ_\ini), \label{eq:tini}
\end{align}
which is advocated in \cite{barfootAssociating2014,long2013banana} to suit particularly well represent odometry errors in terms of pose. Then, ICP estimates the  relative
transformation  between $\bfT_{ \true}$ and $\bfT_{\ini}$, that
is,  outputs an estimate $\hbfT_{\rel}$ of the actual initial error $	\bfT_{\rel}$ which writes
\begin{align}
	\bfT_{\rel}&=\bfT_{\ini}^{-1}\bfT_{\true}=\exp(-\boxi_{\ini})\bfT_{\true}^{-1}\bfT_{\true} \nonumber \\
	&=\exp(-\boxi_{\ini})
\simeq \bfI_4-\boxi_{\ini}^\wedge. \label{eq:trel}
\end{align}

\begin{figure}
	\centering
	\includegraphics{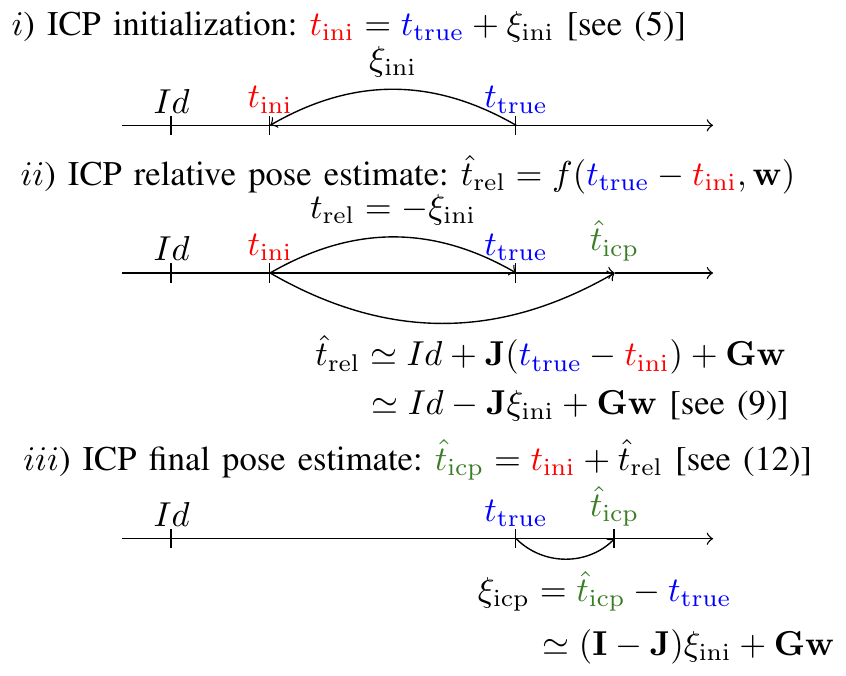}
	\vspace*{-0.5cm}
	\caption{Schematic illustration of the ICP procedure, error definitions and linearizations in the case of 1D translation $t\in \bbR$ ($Id=0$). \label{fig:dessin}\vspace*{-0.3cm}}
\end{figure}

\subsection{ICP  Estimate $\hbfT_{\rel}$   of Relative Pose $\bfT_{\rel}$}

At step $ii$) above,  that is,  once initialization is done, see \cite{pomerleauReview2015}, ICP provides an estimate  $\hbfT_{\rel}$ of  the relative transformation  $	\bfT_{\rel}$ of
\eqref{eq:trel} as a function
\begin{align}
	\hbfT_{\rel} := \icp \left(\calP, \bfT_{\ini}^{-1}\calQ \right) \label{eq:icp}
\end{align}
of the point clouds $\calP$ and $\calQ$. Thus $\hbfT_{\rel}$ appears as a function $f(\bfT_{\rel})\in SE(3)$ of the true relative transformation $\bfT_{\rel}$, typically affected by the phenomena of wrong convergence. Moreover,   sensor (scanner) noise   
induces small fluctuations in the point clouds that affects this estimation.  This yields:
\begin{align}
\hbfT_{\rel}=f(\bfT_{\rel})\exp(
\bfG \bfw) = f\bigl(\exp(-\boxi_{\ini})\bigr)\exp(\bfG 
\bfw)   , \label{eq:sub0}
\end{align}where 
$\bfw\in\mathbb R^{6K}$ encodes errors due to sensor noise on each of the $K$  pairs of  points in the clouds, and $\bfG \bfw\in \mathbb R^6$ the resulting 6 degrees of freedom error made on $\bfT_{\rel}$. Sensor noise  stems from unknown parameters that depend upon the calibration process and drift
with temperature \cite{wangCharacterization2018}.  
If the ICP is
initialized on the true pose $\bfT_\true$ then there is no wrong convergence and the only error stems from noise, i.e.,   $f(\exp(\bfzero) =f(\bfI_4)=\bfI_4$. Thus $f(\cdot) \in SE(3)$ is  close to $\bfI_4$, and   the model may be linearized around $\boxi_{\ini}=\bfzero$, 
$\bfw=\bfzero$ as
\begin{align}
	f\bigl(\exp(-\boxi_{\ini})\bigr)\exp(\bfG
\bfw)&\simeq f(\bfI_4-\boxi_{\ini}^\wedge )\exp(\bfG
\bfw) \nonumber \\ &\simeq \bfI_4 + (-\bfJ\boxi_{\ini} +
\bfG\bfw)^\wedge, \label{eq:l1}
\end{align}
where matrix $\bfJ$  encodes the
linear approximation of $f(\cdot)$.

\subsection{ICP Final Pose Error}\label{sec:icpfinal}

Let us now consider step $iii$) of the ICP algorithm, i.e., the final estimate
\begin{align}
	\hbfT_{\icp}&= \bfT_{\ini} \hbfT_{\rel} \label{10bis:eq}\\
	&= \bfT_\true\exp(\boxi_{\ini})
 f\bigl(\exp(-\boxi_{\ini})\bigr)\exp(\bfG 
\bfw)  .\label{10:eq}
\end{align}
\eqref{10:eq} was obtained substituting   \eqref{eq:tini} and \eqref{eq:sub0} in \eqref{10bis:eq}. Linearizing
\eqref{10:eq} by  recalling \eqref{eq:l1} and keeping only the first order in the small errors
$\boxi_{\ini}$ and $\bfw$ yields $\hbfT_{\icp}  \simeq \bfT_\true\left[ \bfI_4 +
(\boxi_{\ini})^\wedge\right] \left[ \bfI_4 +  \left(-\bfJ \boxi_{\ini} +\bfG\bfw) \right)^\wedge\right] \nonumber
	 \simeq\bfT_\true\left(\bfI_4 + \left(\boxi_{\ini} - \bfJ\boxi_{\ini}+\bfG \bfw\right)^{\wedge}\right) \nonumber  $ and thus in terms of uncertainty representation \eqref{eq:T} we approximately find:
\begin{align}
	\boxed{\hbfT_{\icp}   \simeq \bfT_\true \exp\left((\bfI_6-\bfJ)\boxi_{\ini} + \bfG  \bfw \right).}\label{impfor}
\end{align}
Figure \ref{fig:dessin} recaps the
computations for 1D translations.  There are a couple of   situations of interest. Let us momentarily assume sensor noise to
be turned off, $\bfw=\bfzero$, for simplicity.
\begin{itemize}
	\item If there is one global minimum, then the ICP
	systematically recovers the relative transformation \eqref{eq:trel} at step $ii$) of the algorithm, i.e.
	$\hbfT_\rel=\bfT_{\rel}$ and thus
	$f(\bfT_{\rel} ) = \bfT_{\rel}$. So
	$f(\exp(-\boxi_\ini) )\simeq \bfI_4 - \boxi^\wedge_\ini$ and we identify
	$\bfJ=\bfI_6$ in this case. As a result  the final estimate \eqref{impfor} is $\bfT_\true \exp(\bfzero) = \bfT_\true$ indeed.
\item On the other hand, in the directions where we have no information, e.g.
along  hallways or in underconstrained environment \cite{censiAccurate2007},
the relative transformation will not  be affected in the corresponding directions meaning that (along those directions)
$\bfJ=\bfzero$ and the final error then has the form
$\bfT_\true^{-1}\hbfT_{\icp}=\bfT^{-1}_\true\bfT_\true \exp(\boxi_{\ini}) = \exp(\boxi_{\ini})$, that is, the initialization error fully
remains.
\end{itemize}
In intermediate cases (when there are local minima) the remaining error is a fraction $\bfJ\boxi_{\ini}$ of the initialization error.

\subsection{Corresponding ICP Error Covariance}\label{sec:got}

If we represent ICP uncertainties resorting to concentrated Gaussian
\eqref{eq:T} as
$\hbfT_{\icp}\sim\calN_L\left(\bfT_\true, \bfQ_{\icp}\right)$, i.e., we posit $\bfT_\true^{-1}\hbfT_{\icp}=\exp(\boxi_{\icp})$, then the covariance matrix $\bfQ_{\icp}$ of $\boxi_{\icp}$ describes dispersion (hence uncertainty) of the ICP error. Plugging the latter representation into \eqref{impfor}, we  have
$\boxi_{\rmicp} = (\bfI_6-\bfJ)\boxi_{\ini} + \bfG \bfw$. As
the initialization error is assumed to have covariance matrix $\bfQ_{\ini}$
 typically inferred though an odometry error model \cite{barfootAssociating2014,long2013banana} and by denoting $\bfQ_{\mathrm{sensor}}$ the covariance of scan sensor noise
$\bfw$, the covariances add up owing to independence of sensor noises, and by squaring $\boxi_{\rmicp} = (\bfI_6-\bfJ)\boxi_{\ini} + \bfG \bfw$ we find
\begin{align}\boxed{
	\bfQ_{\icp} = (\bfI_6-\bfJ)\bfQ_{\ini}(\bfI_6-\bfJ)^T + \bfG\bfQ_{\mathrm{sensor}}\bfG^T. }\label{eq:Qicp}
\end{align}
This is our first result about ICP covariance. The first term related to the initialization uncertainty and that accounts for
wrong convergence, lack of constraints in the clouds  and unobservable directions, and the second one related
to scan noise and that may be computed through ``Censi-like" \cite{censiAccurate2007}
formulas as we will show  in section \ref{sec:cova}.

\subsection{Discussion}
Albeit not
obvious, $\bfJ$ actually heavily depends on $\bfQ_{\ini}$. This is an  insight of the present paper: \emph{uncertainty
of ICP does not exist in itself}. 
Assume indeed there are various local minima. If $ \bfQ_{\ini}$ is very
small, then  all initializations $\bfT_{ \ini}$ fall within the attraction basin
of $\bfT_{\true}$ and thus $f(\bfT_{\rel})=\bfT_{\rel}$ and we identify
$\bfJ=\bfI_6$. But if $ \bfQ_{\ini}$ is large enough only a fraction of
initializations  $\bfT_{ \ini}$  lead to $f(\bfT_{\rel})=\bfT_{\rel}$, the ones that get
trapped in other local minimas do not lead to correct estimate of $\bfT_{\rel}$
and  $\bfJ\neq\bfI_6$. Thus $\bfJ$ is not the analytical Jacobian of function
$f(\cdot)$, and may be viewed as   its  ``statistical
linearization''  \cite{gustafssonSome2012}. This prompts the use of an unscented transform  \cite{juliernew2000} to compute it, see Section \ref{sec:prop_T_init}.

\subsection{Maximum Likelihood Fusion of Initial and ICP Estimates}
$\bfT_{\ini}$ and  $\hbfT_{\icp}$ may be viewed as two estimates of
$\bfT_{\true}$ associated with uncertainty respectively
$\bfQ_{\ini}=\mathrm{cov}(\boxi_{ \ini})$ and
$\bfQ_{\icp}=\mathrm{cov}(\boxi_{\rmicp})$ where $\boxi_{\rmicp} = (\bfI
-\bfJ)\boxi_{\ini} + \bfG \bfw$. The corresponding  pose fusion problem of
finding the Maximum Likelihood (ML) of a pose  $\bfT_{\true}$ given two
uncertain pose estimates was considered in  \cite{barfootAssociating2014}, with
the important difference that herein $\boxi_{\ini} $ and $\boxi_{\icp} $ are not
independent, they are \emph{correlated}, with joint  matrix of initialization and
ICP errors
    \begin{align}
	\bfQ:=\mathrm{cov}\bigl(\begin{bmatrix}\boxi_{ \ini}\\\boxi_{\rmicp}\end{bmatrix}\bigr)=  \begin{bmatrix}\bfQ_{\ini}&\bfQ_{\ini}(\bfI-\bfJ)^T\\ (\bfI-\bfJ)\bfQ_{\ini}&\bfQ_{\icp}\end{bmatrix}.\label{allright4}
	\end{align}
Using linearization as previously and following first-order computations in \cite{barfootAssociating2014}, the maximum likelihood estimate of $\bfT_{\true}$ may be approximated   as $\hbfT_{\mathrm{ML}}=\bfT_{\true}\exp(\boxi_{ \mathrm{ML}})$ with $\mathrm{cov}(\boxi_{ \mathrm{ML}})=\bfQ_{ \mathrm{ML}} $, with  $\bfQ_{ \mathrm{ML}} $  defined through its inverse:
 \begin{align}
\bfQ_{ \mathrm{ML}}^{-1}=\begin{bmatrix}\bfI\\\bfI\end{bmatrix}^T\begin{bmatrix}\bfQ_{\ini}&\bfQ_{\ini}(\bfI-\bfJ)^T\\ (\bfI-\bfJ)\bfQ_{\ini}&\bfQ_{\icp}\end{bmatrix}^{-1}\begin{bmatrix}\bfI\\\bfI\end{bmatrix}.\label{eq:Qml}
\end{align}
{The latter stems from classical linear estimation theory and may be proved
using the Kalman information filter:   to the first order  $\hbfT_{\ini}$ and
$\hbfT_{\icp}$ are considered as two noisy measurements of $\bfT_{\true}$ with
joint covariance \eqref{allright4}, the measurement matrix is thus $\bfH:=
\begin{bmatrix}\bfI&\bfI\end{bmatrix}^T$ and as    $\bfT_{\true}$  is initially
totally unknown  the  prior covariance   satisfies $\bfP^{-1}=\bfzero$. The
covariance of the Kalman estimate in the light of measurements is thus updated
in information form as $\bfP^{-1}\leftarrow\bfzero+\bfH^T\bfQ^{-1}\bfH.$}

\section{Practical Covariance Computation}\label{sec:cova}

This section describes our algorithm for estimating the 3D ICP uncertainty
covariance \eqref{eq:Qicp} leveraging  findings of Section \ref{sec:source}. We propose to  first compute
the rightmost term of \eqref{eq:Qicp} which is due to sensor noise.

\subsection{Computation of Dispersion Owing to Sensor Noise}\label{sec:silvere}We now focus on the computation of
  $\bfG\bfQ_{\mathrm{sensor}}\bfG^T$.  The cost function of point-to-plane   ICP after initialization writes $J_{\hbfT_{\rel}} \left(\calP, \bfT_{\ini}^{-1}\calQ \right) =\sum_{k=1}^{K} \|\bigl(\hbfT_{\rel}p_k -\tilde q_k\bigr)\cdot n_k \|^2$, where the $\tilde q_k$'s denote the points of  $\bfT_{\ini}^{-1}\calQ$ and  $K$ is the number of pairs of matched points. Linearizing on $SE(3)$,  we may linearize the cost $ J_{\hbfT_{\rel}\exp\left(\boxi\right)} \left(\calP, \bfT_{\ini}^{-1}\calQ \right) = \sum_{k=1}^{K} \|\bigl(\hbfT_{\rel}\exp\left(\boxi\right)p_k -\tilde q_k\bigr)\cdot n_k \|^2  $ w.r.t. estimate $\hbfT_{\rel}$ at $\hbfT_{\rel}=\bfT_{\rel}$  as
\begin{align}
  J_{\hbfT_{\rel}\exp\left(\boxi\right)} \left(\calP, \bfT_{\ini}^{-1}\calQ \right)  
 \simeq \sum_{k=1}^{K} \|\bfB_k \boxi   - d_k \|^2, \label{eq:lin}
\end{align}with 
$d_k$ a scalar being function of
differences between pairs of points and point normals. Least squares formulas yield an optimal value  $\boxi^*= \bfA^{-1}  \sum_{k=1}^{K} \bfB_k^T d_k$, where we let $\bfA = \sum_{k=1}^{K} \bfB_k^T \bfB_k$. Each $d_k$ is affected by $k$-th component $w_i$ of previously introduced  sensor noise $\bfw$, and this induces fluctuations in  $\boxi^*$ over various experiments. Let's postulate  $w_i=\bfb+\nu_i$ with $\nu_i$ a white noise of variance $\sigma^2$, and $\bfb$ and \emph{unknown} calibration bias that is identical for all points but varies from one experiment to the next. Following least
squares covariance, see \cite{censiAccurate2007, bonnabelCovariance2016}, we end up with: 
\begin{align}
	\bfG\bfQ_{\mathrm{sensor}}\bfG^T &=\sigma^2 \bfA^{-1} + \bfA^{-1} \bfB \cov\left(\bfb\right) \bfB^T \bfA^{-1}, \label{eq:Q1}
\end{align}
where $\bfA = \sum_{k=1}^{K} \bfB_k^T \bfB_k$, and $\bfB =\sum_{k=1}^{K} \bfB_k^T $.
We recover the covariance $\sigma^2 \bfA^{-1}$ of \cite{censiAccurate2007,
bonnabelCovariance2016} w.r.t. sensor white noise, and a new term, $\bfA^{-1}
\bfB \cov\left(\bfb\right) \bfB^T \bfA^{-1}$, that represents the covariance
w.r.t. the unknown bias $\bfb$, that is, correlated noise. This new additional
term is paramount as  $\bfA$ has magnitude proportional to   $K$, hence $\bfA^{-1}$ is very small,
explaining that Censi's formula (based on $\bfA^{-1}$ only) seems
overoptimistic  \cite{mendesICPbased2016}. For example $\bfA^{-1}$ has trace \SI{0.2}{cm}  for the registration displayed in
Figure \ref{fig:Q_init} whereas the covariance
w.r.t. the unknown bias has trace \SI{2.6}{cm}. In practice  $\bfb$ arises from
sensor calibration, laser stability \cite{wangCharacterization2018},
observed material \cite{pomerleauNoise2012}, and  incidence of beams
\cite{laconteLidar2019}. In the remainder we assume bias standard deviation  to be approximately    \SI{5}{cm} as  in
\cite{pomerleauNoise2012}.

Note that  \eqref{eq:Q1} captures the effect of underconstrained situations like hallways. Indeed in  unobservable directions the cost $J_{\hbfT_{\rel}}(\cdot)$ is constant, yielding  small eigenvalues for $\bfA$ and hence large covariance \eqref{eq:Q1}.
Derivation of $\bfB_k$   and extraction of
$\bfG$ in \eqref{eq:Q1} are  available with
paper code.

\begin{algorithm}[t]
\KwIn{$\calP, \calQ, \bfT_{\ini}, \bfQ_{\ini}, \hbfT_{\rmicp}=\bfT_{\ini} \icp\left(\calP,  \bfT_{\ini}  ^{-1}
\calQ\right)$\;}
\tcp{set sigma points}
\nl $\boxi^j_{\ini} =  \col\left(\sqrt{6 \bfQ_{\ini}} \right)_{j}, ~j=1,\ldots,6$,\\
$\boxi^j_{\ini} =  - \col\left(\sqrt{6 \bfQ_{\ini}} \right)_{j-6}, ~j=7,\ldots,12$\;
\tcp{propagate sigma points through (7)}
\nl$\bfT_{\ini}^{j} = \bfT_{\ini} \exp\left(\boxi_{\ini}^j\right), ~j=1,\ldots,12$\;
 $\hbfT_{\rmicp}^j= \bfT_{\ini}^{j} \icp\left(\calP, (\bfT_{\ini}^{j})^{-1}
\calQ\right), ~j=1,\ldots,12$\;
\nl $\hboxi_{\rmicp}^j = \exp^{-1}\left(\hbfT_{\rmicp}^{-1} \hbfT^{j}_{\rmicp}\right), ~j=1,\ldots,12$\;
\tcp{compute covariance and infer $\bfJ$}
\nl $(\bfI_6-\bfJ)\bfQ_{\ini}(\bfI_6-\bfJ)^T = \sum_{j=1}^{12} \frac{1}{12} \hboxi_{\rmicp}^j \hboxi_{\rmicp}^{jT}$\;
\nl $\hboxi_{\rmicp} = \sum_{j=1}^{12}\frac{1}{12} \hboxi_{\rmicp}^j$\;
\nl $\bfJ = -\left(\sum_{j=1}^{12} \frac{1}{12} \left(\hboxi_{\rmicp}^j
- \hboxi_{\rmicp}\right) \boxi_{\ini}^{jT}\right) \bfQ_{\ini}^{-1} + \bfI_6$\;
\KwOut{$\bfJ,~(\bfI_6-\bfJ)\bfQ_{\ini}(\bfI_6-\bfJ)^T$\;}
\caption{Computation of matrix $\bfJ$ in \eqref{eq:Qicp}\label{alg:1}}
\end{algorithm}

\subsection{Computation of  Dispersion owing to ICP Initialization}\label{sec:prop_T_init}
Computation of $(\bfI_6-\bfJ)\bfQ_{\ini}(\bfI_6-\bfJ)^T$ in \eqref{eq:Qicp} is of greater importance as in practice it largely dominates $\bfG\bfQ_{\mathrm{sensor}}\bfG^T$. We propose to compute it in a deterministic derivative-free method, in which we adapt the unscented transform \cite{juliernew2000} for pose by following  \cite{brossardUnscented2017,barfootAssociating2014}. The advantages of using our unscented based method rather than Monte-Carlo sampling are fourfold: 1) it is deterministic; 2) it remains computationally reasonable by adding only 12 ICP registrations; 3) it explicitly computes the cross-covariance matrix between $\hbfT_{\rmicp}$ and $\bfT_{\ini}$ as a by-product without extra computational operations; and 4) it scales with $\bfQ_{\ini}$, i.e. our approach naturally self-adapts to the confidence we have in initialization without extra parameter tuning.

We compute the covariance as follows, see Algorithm \ref{alg:1}:
\begin{itemize}
	\item we consider the prior distribution $\bfT_{\mathrm{prior}} \sim \calN_L(\bfT_{\ini}, \bfQ_{\ini})$, which is approximated by a set of so-called sigma-points $\boxi_\ini^j$, see step 1);
	\item we approximate the propagated distribution $\bfT_{\mathrm{prop}} = \bfT_{\mathrm{prior}} \icp (\calP, \bfT_{\mathrm{prior}}^{-1}\calQ)$ as
	\begin{align}
	\bfT_{\mathrm{prop}} &= \calN_L\left(\bfT_{\ini}, \bfQ_{\ini}\right) \icp \left(\calP, \calN_L\left(\bfT_{\ini}, \bfQ_{\ini}\right)^{-1}\calQ\right) \nonumber \\
	&\sim \calN_L\left(\hbfT_{\rmicp}, (\bfI_6-\bfJ)\bfQ_{\ini}(\bfI_6-\bfJ)^T\right),
	\end{align}
after propagating each sigma-point in steps 2) and 3), where $\hbfT_{\rmicp}$ is
the given ICP pose estimate.
We compute the covariance and  infer the matrix  $\bfJ$ as a by-product in respectively steps 4) and
6).
\end{itemize}

\renewcommand{\figurename}{Table}
\setcounter{figure}{0} 
\begin{figure}
	\centering
	\scriptsize
	\begin{tabular}{c||c|c|c|c||c|c|c|c}
		\toprule
\multirow{2}{*}{\scriptsize metric}&  \multicolumn{2}{c|}{ NNE}  & \multicolumn{2}{c||}{ KL
div.} &  \multicolumn{2}{c|}{ NNE*}  & \multicolumn{2}{c}{ KL div.*}   \\
		&   trans. &  rot. &  trans.  &  rot. &   trans. &  rot. &  trans.  &  rot.\\
		\midrule
		 $\hbfQ_{\mathrm{censi}}$ & \tiny\num{e3} &  \tiny\num{e3} &  \tiny\num{e4}&  \tiny\num{e5} & \tiny\num{38} &  \tiny\num{e2} &  \tiny\num{e3}&  \tiny\num{e5}\\
$\hbfQ_{\mathrm{carlo}}^{\mathrm{monte}}$ & \tiny\num{e3} &  \tiny\num{e2} &  \tiny\num{e4}&
\tiny\num{e4} & \tiny\num{22} &  \tiny\num{20} &  \tiny\num{e3}&  \tiny\num{e3}\\
\textbf{proposed} & \textbf{\tiny\num{4.2}} & \textbf{\tiny\num{34}} &\textbf{\tiny\num{e2}}&
\textbf{\tiny\num{e2}} & \textbf{\tiny\num{0.8}} & \textbf{\tiny\num{3.8}} &\textbf{\tiny\num{31}}& \textbf{\tiny\num{98}}\\
		\bottomrule
	\end{tabular}
\caption{Results of ICP uncertainty estimation in term of Normalized Norm Error
(NNE) and Kullbach-Leibler divergence (KL div.) divided into translation and
rotation parts.
As the ICP error distributions are not
Gaussian \cite{pomerleauComparing2013}, we provide robust statistics
(starred, *) by
removing both the more and less accurate quantiles of each registration.
The proposed method outperforms the two others. \label{fig:results2}} \vspace*{-0.1cm}
\end{figure}
\renewcommand{\figurename}{Fig.}
\setcounter{figure}{2} 

We derive the algorithm by following \cite{brossardUnscented2017} for pose
measurement, zero-mean prior distribution, and where we set $\alpha=1$.

\section{Experimental Results}\label{sec:results}
\subsection{Dataset Description \& ICP Algorithm Setting}\label{sec:dataset}
This section evaluates the ability of the approach to estimate   ICP uncertainty
on the \emph{Challenging data sets for point cloud registration	algorithms}
\cite{pomerleauChallenging2012}. It comprises eight sequences where point clouds
are taken in environments ranging from structured to unstructured, and indoor to
outdoor. Each sequence contains between 31 and 45 point cloud scans along with
ground-truth pose for each scan, that provides a total of \num{268} scans
and \num{1020} different registrations as we align each scan with the three scans the following.

We configure the ICP as in \cite{pomerleauComparing2013} with 95\% random sub-sampling, kd-tree for data association, and point-to-plane error metric where we keep the 70\% closest point associations for rejecting outliers.

\subsection{Compared Methods and Evaluation Metrics}

This section evaluates the following methods:
\begin{itemize}
	\item[$\hbfQ_{\mathrm{censi}}$]: the close-form method of \cite{censiAccurate2007} adapted for the ICP setting defined above;
	\item[$\hbfQ_{\mathrm{carlo}}^{\mathrm{monte}}$]: the covariance computed after sampling of 65 Monte-Carlo ICP estimates;
	\item[$\hbfQ_{\icp}$]: our proposed approach detailed in Section \ref{sec:cova}.
\end{itemize}
Each method assumes depth sensor white noise and bias with \SI{5}{cm} standard deviation, which is the mean value found in \cite{pomerleauNoise2012} for the Hokuyo sensor used for these experiments, and all methods know the initial uncertainty $\bfQ_{\ini}$, whose magnitude \SI{0.1}{m} and \SI{10}{\deg} corresponds to the easy scenario of \cite{pomerleauComparing2013}.

\begin{figure}
	\includegraphics{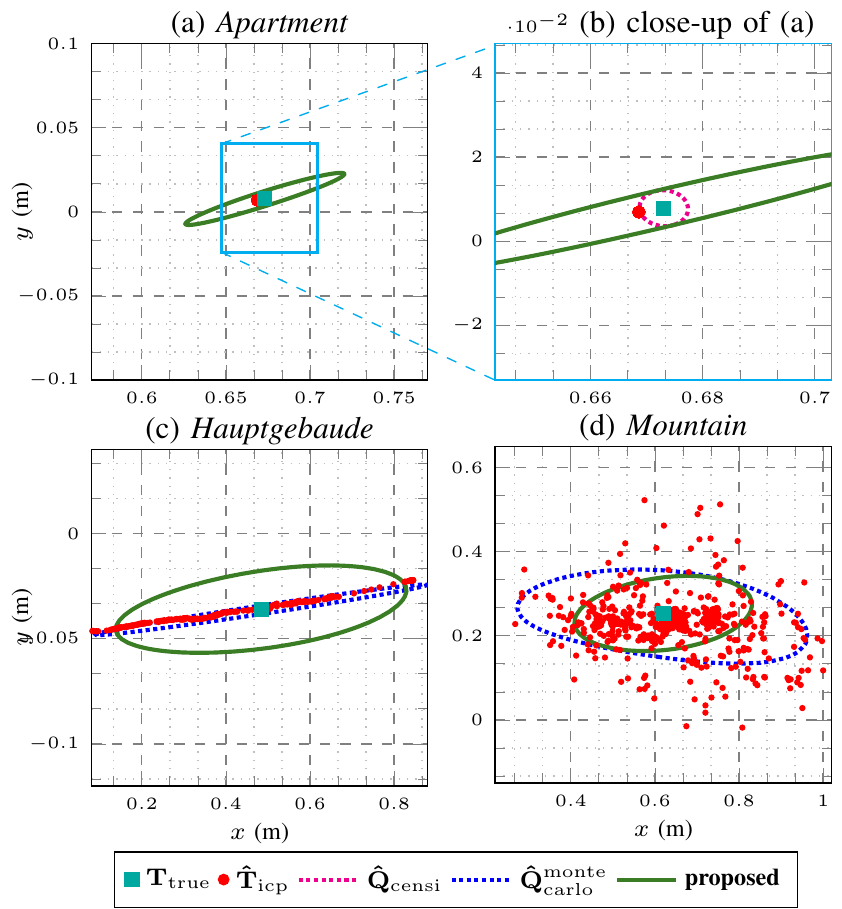}
	\vspace*{-0.0cm}
	\caption{\label{fig:results}Results on real data of \cite{pomerleauChallenging2012} projected onto the ground plane for visualization.   Ellipses represent the 95\% (3$\sigma$) confidence sets for each uncertainty estimation method.  (a): ``true convergence situation", the errors are mainly caused by sensor noises and Censi's formula should apply. (b): however we see the Censi ellipse seems optimistic as ground truth is almost outside it whereas it falls well within our ellipse (green). (c, d): ``wrong convergence", the large errors are due ICP initialization. Only our approach is consistent with ICP uncertainty in each environment and does not suffer from overoptimism.\vspace*{-0.3cm}}
	\end{figure}

	\renewcommand{\figurename}{Table}
	\setcounter{figure}{1} 
	\begin{figure*}
		\centering
		\scriptsize
		\begin{tabular}{c||c|c||c|c||c|c||c|c||c|c||c|c||c|c||c|c}
			\toprule
			sequence & \multicolumn{2}{c||}{\emph{Apartment}} & \multicolumn{2}{c||}{\emph{Hauptgebaude}} & \multicolumn{2}{c||}{\emph{Stairs}} & \multicolumn{2}{c||}{\emph{Mountain}}& \multicolumn{2}{c||}{\emph{Gazebo summer}} & \multicolumn{2}{c||}{\emph{Gazebo winter}} & \multicolumn{2}{c||}{\emph{Wood summer}} & \multicolumn{2}{c}{\emph{Wood winter}}\\
			\midrule
			Mah. dist. & \scriptsize trans.  & \scriptsize rot. & \scriptsize trans.  &\scriptsize rot. & \scriptsize trans.  &\scriptsize rot. &\scriptsize  trans.  &\scriptsize rot.&  \scriptsize trans.  & \scriptsize rot. & \scriptsize trans.  & \scriptsize rot.& \scriptsize trans.  & \scriptsize rot. & \scriptsize  trans.  & \scriptsize rot.\\
			\midrule

			CELLO-3D\cite{landryCELLO3D2019} &\num{0.2}&\num{0.1}&\num{0.3}&\num{0.2}&\num{0.1}&\num{0.2}&-&-&\num{0.2}&\num{0.2}&\num{0.1}&\num{0.2}&\num{0.1}&\textbf{\num{0.3}}&\num{0.1}&\textbf{\num{0.3}}\\
			ini.+ICP & \num{3.5}&\num{15}&\num{1.9}&\num{3.2}&\textbf{\num{1.1}}&\textbf{\num{4.2}}&\num{1.5}&\textbf{\num{1.2}}&\num{1.1}&\textbf{\num{2.1}}&\num{1.9}&\textbf{\num{3.7}}&\textbf{\num{1.5}}&\num{4.6}&\textbf{\num{1.2}}&\num{4.8}\\
			\textbf{proposed} &\textbf{\num{2.3}}&\textbf{\num{9.8}}&\textbf{\num{1.8}}&\textbf{\num{2.9}}&\textbf{\num{1.1}}&\textbf{\num{4.2}}&\textbf{\num{1.2}}&\textbf{\num{1.2}}&\textbf{\num{1.0}}&\num{2.3}&\textbf{\num{1.8}}&\textbf{\num{3.7}}&\textbf{\num{1.5}}&\num{4.7}&\textbf{\num{1.2}}&\num{4.2}\\
			\bottomrule
		\end{tabular}

		\caption{Trajectory consistency results in term of Mahalanobis distance (bold indicates best performance) split into translation and rotation parts for the sequences of \cite{pomerleauChallenging2012}, where \emph{Mountain} is not considered in \cite{landryCELLO3D2019}. Our method obtains on average the best uncertainty assessment, albeit slightly optimistic. \label{fig:results_graph2}\vspace*{-0.4cm}}
	\end{figure*}
	\renewcommand{\figurename}{Fig.}
	\setcounter{figure}{3} 

We compare the above methods using two metrics:
\subsubsection{Normalized Norm Error (NNE)} that evaluates the historically challenging  \cite{censiAccurate2007, bengtssonRobot2003} prediction of the covariance scale, and is	computed as
\begin{align}
\mathrm{NNE} = \big(\frac{1}{N}\sum_{n=1}^N \|\boxi_n\|_2^2/\trace(\hbfQ_n)\big)^{1/2}, \label{eq:nne}
\end{align}
where $\boxi_n = \exp^{-1}(\bfT_\true^{-1} \hbfT_{n})$ with  is the transformation error and $\hbfQ_n$ the estimated uncertainty covariance matrix, and averaged over $N$ samples. This metric  characterizes the uncertainty as only the true registration is known (the exact distribution of the point cloud is unknown). The target value is one, below one the estimation is pessimistic, whereas a value over one indicates an overoptimistic estimation.
\subsubsection{Kullback-Leibler Divergence (KL div.)} which is computed between a pseudo-true distribution and the estimated distribution. The pseudo-true distribution is computed after sampling \num{1000} ICP estimates of the evaluated registration over the initial position. As sensor noise is fixed in the point clouds, this distribution represents the uncertainty stemming from  initialization errors.

\subsection{Results}\label{sec:cov_results}

Results are averaged over \num{1000} initializations for each of the \num{1020}
considered pairs of point clouds, representing a total of more than one million registrations,
where the ICP is initialized with a different estimate $\bfT_{\ini}$ sampled
from $\calN_L(\bfT_\true, \bfQ_{\ini})$. Table
\ref{fig:results2} provides average results over the eight sequences, and Figure \ref{fig:results}
illustrates typical registrations from structured to unstructured environments.
We observe:
\begin{itemize}
	\item $\hbfQ_{\mathrm{censi}}$ is far too optimistic and unreliable for sensor-fusion, as noted in \cite{mendesICPbased2016}. Its centimetric confidence interval makes sense only when ICP is very accurate;
\item $\hbfQ_{\mathrm{carlo}}^{\mathrm{monte}}$ is overoptimistic when the
discrepancy arising from ICP initialization remains negligible, see Figure
\ref{fig:results} (b), for which the method predicts a confidence interval
with millimetric size. This is naturally explained as the method assumes no
error caused by sensor noises;
\item the proposed method obtains the best results for both  metrics as
displayed in  Table \ref{fig:results2}. It notably outperforms
$\hbfQ_{\mathrm{carlo}}^{\mathrm{monte}}$ while deterministic hence more
reliable, and computationally much cheaper. The dominant term is generally due to initial uncertainty. However in ``global minimum" cases    the sensor bias
 used for computing $\hbfQ_{\rmicp}$ slightly inflates the
covariance of \cite{censiAccurate2007}, and more closely
captures actual uncertainty, see Figure
\ref{fig:results} (a,b).
\end{itemize}
Besides outperforming the other methods, our method provides simple  parameter
tuning: we set the bias noise standard deviation as having same magnitude as
sensor white noise, and the error stemming from ICP initialization does not need
to be tuned when $\bfQ_{\ini}$  is an output of inertial, visual, or wheeled
odometry system  \cite{genevaLIPS2018, dubeOnline2017}.

Regarding computational complexity and execution  time, step 2) of the algorithm requires 12 registrations which take \SI{6}{s} when registrations are  computed parallely, whereas the remaining part of the algorithm takes less than \SI{0.1}{s}. The 65 Monte-Carlo runs are more than five times more demanding than the the proposed method.

\section{Complementary Experimental Results}
\label{sec:results2}
This section provides an in-depth analysis of the method in term of trajectory
consistency, robustness to high and misknown initial uncertainty, and
discusses  the advantages and the  validity of the approach.

\subsection{Application to Trajectory Consistency}

We asses the quality of the covariance estimation in Section \ref{sec:cova} over
trajectories as follows. For each sequence of \cite{pomerleauChallenging2012}, we
compute the global pose estimate at
scan $l$ by compounding transformations such that $\hbfT^l = \hbfT^{0,1} \ldots
\hbfT^{l-1,l}$, whose covariance is computed with the
closed-form expressions of \cite{barfootAssociating2014} which
are valid up to 4-th order approximation. We compare three methods defined as:
\begin{itemize}
	\item[CELLO-3D]: reproduced results  of \cite{landryCELLO3D2019},  that proposes a learning based method for estimating the ICP covariance, which is trained on environments similar  to the tested sequence. The results are indicative as the ICP setting of \cite{landryCELLO3D2019} slightly differs from the setting of \cite{pomerleauComparing2013} we use;
\item[ini.+ICP]:  combines initialization and ICP measurements with the covariance
estimate \eqref{eq:Qml} without considering cross-covariance terms, i.e., applying formulas of  \cite{barfootAssociating2014} ;
\item[\textbf{proposed}]: based on the full proposed covariance of the maximum-likelihood estimate \eqref{eq:Qml}.
\end{itemize}
We set initial errors as in Section \ref{sec:results} and evaluate the above methods using the \emph{Mahalanobis Distance}
  proposed in \cite{landryCELLO3D2019} between  final trajectory estimates and   ground truth
	\begin{align}
	\text{Mah. dist.} = \Big(\sum_{n=0}^N \frac{\boxi_n^T \hbfQ_n^{-1} \boxi_n}{\dim(\boxi_n)N} \Big)^{1/2},
	\end{align}
where $\boxi_n = \exp^{-1}(\bfT_\true^{-1} \hbfT_{n})$ is the transformation error and
$\hbfQ_n$ the estimated covariance matrix,  averaged over $N$
samples. The target value is one, below one the estimation is
pessimistic, and above one  the estimates are  optimistic.

\begin{figure}
	\centering
	\includegraphics{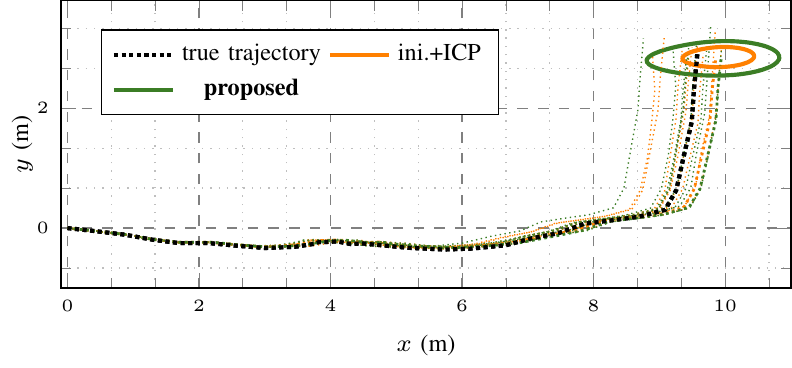}
	\vspace*{-0.1cm}
	\caption{\label{fig:pose-graph} Results projected onto the ground plane for visualization in the  \emph{Stairs} sequence of \cite{pomerleauChallenging2012}, where the ``ellipses" (lines) represent the 95\% ($3\sigma$) final confidence sets.\vspace*{-0.5cm}}
\end{figure}

We average results over 40 different initial trajectories for each sequence,
which are numerically displayed in Table \ref{fig:results_graph2} and illustrated in Figure \ref{fig:pose-graph}. We observe:
\begin{itemize}
	\item CELLO-3D is the only pessimistic method, which estimates uncertainty ranging from 3 to  10 times higher than actual uncertainty. It evidences how difficult it is to asses ICP uncertainty in practice;
	\item the proposed approach obtains on average the best results. It obtains similar estimates than ini.+ICP when the
	ICP algorithm is accurate (\emph{Gazebo} and \emph{Wood}).
	In more difficult environments, e.g. \emph{Stairs}, it better incorporates initialization than ini.+ICP thanks to it
	accounting for measurement correlation encoded in \eqref{allright4}, see Figure \ref{fig:pose-graph}.
\end{itemize}

These results confirm the ability of the method to compute covariance estimates over trajectories also and the relevance of correlation terms between ICP and initial estimates.

\subsection{Role of Initial Uncertainties in Covariance Estimation}\label{sec:inf}

\begin{figure}
	\centering
	\includegraphics{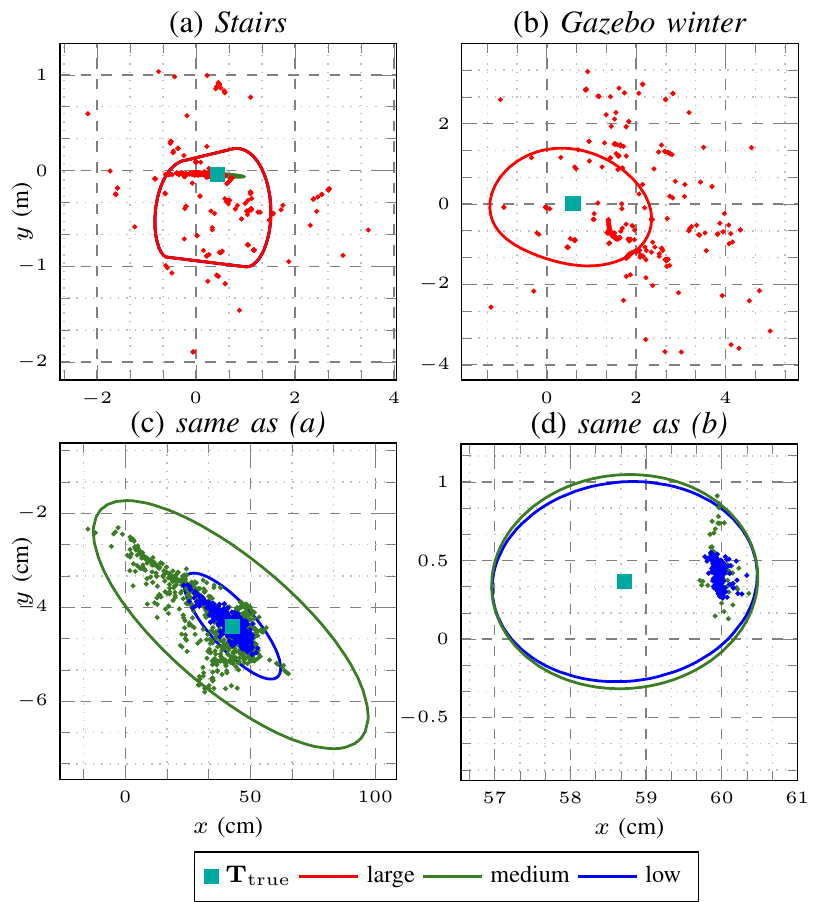}
	\vspace*{-0.1cm}
	\caption{\label{fig:7}ICP results (dots) and 95\% ($3\sigma$) confidence sets (lines) following our method for three levels of initial uncertainty. We see the latter highly influences the ICP registrations and ICP covariance estimations accordingly.}
	\end{figure}

We evaluate the influence of $\bfQ_\ini$ on the
covariance estimation in challenging situations where $\bfQ_\ini$ is high,
inaccurately known (the estimation of $\bfQ_\ini$ is in itself  challenging), and sensor noise is inflated.
For each environment of \cite{pomerleauChallenging2012}, we evaluate the
method in 9 situations where initial uncertainty is easy, medium
and difficult with
respectively \num{0.1}, \num{0.5}, \SI{1}{m} and \num{10}, \num{20},
\SI{50}{deg} standard deviation, see \cite{pomerleauComparing2013}. In each situation we evaluate the algorithm with different magnitudes for $\bfQ_\ini$, hence assessing its robustness to pessimistic and optimisitic parametrization. We
finally add white and depth bias noises on already noisy point clouds with
\SI{5}{cm} standard deviation.
Results are given in Table \ref{fig:com}, and illustrated in
Figure \ref{fig:7}.
\begin{itemize}
\item The ICP algorithm obtains unreliable results for large initial uncertainty, see
Figure \ref{fig:7}, whereas it obtains centimetric errors for low levels of
initial uncertainty. As anticipated in Section
\ref{sec:rol}, significant ICP errors are
caused by inaccurate initialization;
\item  ICP final outputs are agnostic to
initialization when a global
minimum exists, see e.g. the \emph{Gazebo} and \emph{Wood}
environments in Table \ref{fig:com} for   levels of $\bfQ_\ini$ corresponding to easy and medium scenario. The   method obtains   correct
estimates where the sensor noise terms numerically dominate the estimated
covariances, and  $\bfJ\approx \bfI$, see Section \ref{sec:icpfinal};
\item another environments contain local minima, e.g. \emph{Hauptgebaude}. Then the
algorithm outputs reflect the pessimistic or optimistic belief about
the initial uncertainty. We recommend in these situations to set $\bfQ_\ini$
sufficiently high to favour conservatism;
\item our method is able to detect inaccurate ICP registrations by providing
very high covariance estimates, although it cannot accurately describe non Gaussian distributions, see Figure \ref{fig:7} (a,b).
\end{itemize}

\subsection{Discussion about the Proposed Approach}

We finally examine the pros, cons, and fundamental assumptions of the method. The main
advantages are:  it being anchored in a mathematical theory, its efficiency to assess uncertainty for
acceptable levels of initial uncertainties, its simplicity, while being
computationally reasonable, see Algorithm 1. The cross-covariance term in
\eqref{eq:Qml} may   be fruitful for increasing robustness of back-end systems, e.g.
pose-graph \cite{mendesICPbased2016}, as it correlates
two previously supposed independent measurements. Comparisons between diagonal terms in
\eqref{eq:Qml} finally provides a way for trading-off between initial odometry
guesses and ICP estimates, see \cite{ovchinnikovWindowed2019}.

The Gaussian error assumption of the ICP estimates is the core hypothesis of the
method. We required this assumption to obtain a tractable method able to provide
a covariance for a state estimator, e.g. a Kalman filter.   However, if one pursues a more accurate estimation of the ICP distribution, we suggest massive sampling methods
as an expansive alternative, although our method largely proves sufficient to detect problematic situations.

The method finally requires the covariance of the initial uncertainty as an input. If the
provided initialization confidence is inexact,
the method outputs may reflect the initial  optimism or pessimism in general situations. Nonetheless
an insight of the present paper, see Section \ref{sec:source} and e.g. Figure \ref{fig:7}, is that ICP errors intrinsically depend on
the initial accuracy so that a coarse idea of initial uncertainty is essential,
whatever the method one desires to use.

Finally, \cite{long2013banana,barfootAssociating2014,bourmaudContinuousDiscrete2015} show that
concentrated Gaussian distribution \eqref{eq:T} faithfully describe robot
odometry models, such that methods like the Kalman filter are able to provide an
accurate (concentrated) Gaussian approximation to the true initial uncertainty
for which our approach has been designed.

\renewcommand{\figurename}{Table}
\setcounter{figure}{2} 
\begin{figure}
	\centering
	\scriptsize
	\begin{tabular}{c||c|c|c|c|c|c|c|c|c}
		\toprule
		\tiny$\bfQ_\ini$(true)&  \multicolumn{3}{c|}{easy}  & \multicolumn{3}{c|}{medium} &
\multicolumn{3}{c}{difficult} \\
\midrule
\tiny$\hbfQ_\ini$(algo)& \tiny easy  &\tiny med. &\tiny diff.&\tiny easy  & \tiny med.
&\tiny diff.&\tiny easy  &\tiny med. &\tiny diff. \\
		\midrule
		Appart. &\tiny\num{1.6} & \tiny\num{1.5} & \tiny$10^{\text{-}6}$ & \tiny\num{1.8} & \tiny\num{1.7} &\tiny$10^{\text{-}6}$& \tiny\num{e5} & \tiny\num{e5} & \tiny\num{0.6} \\
		Haupt. &\tiny\num{1.8} & \tiny\num{0.3} & \tiny$10^{\text{-}3}$ & \tiny\num{3.6} & \tiny\num{0.7} & \tiny$10^{\text{-}3}$ & \tiny\num{e3} & \tiny\num{e3} & \tiny\num{0.3} \\
		Stairs &\tiny\num{0.7} & \tiny\num{0.2} & \tiny$10^{\text{-}3}$ & \tiny\num{1.9} & \tiny\num{0.8} & \tiny$10^{\text{-}3}$ & \tiny\num{e3} & \tiny\num{e3} & \tiny\num{0.7} \\
		 Montain &\tiny\num{2.1} & \tiny\num{1.1} & \tiny$10^{\text{-}3}$ & \tiny\num{3.3} & \tiny\num{1.8} & \tiny$10^{\text{-}3}$ & \tiny\num{e3} & \tiny\num{e3} & \tiny\num{4.0} \\
		 Gazebo &\tiny\num{1.1} & \tiny\num{1.0} & \tiny$10^{\text{-}5}$ & \tiny\num{1.3} & \tiny\num{1.2} & \tiny$10^{\text{-}5}$ & \tiny\num{e4} & \tiny\num{e4} & \tiny\num{0.8} \\
		 Wood &\tiny\num{2.1} & \tiny\num{2.1} & \tiny$10^{\text{-}3}$ & \tiny\num{2.2} & \tiny\num{2.2} & \tiny$10^{\text{-}3}$ & \tiny\num{e3} & \tiny\num{e3} & \tiny\num{0.3} \\
		\bottomrule
	\end{tabular}
	\caption{\label{fig:com} NNE, see \eqref{eq:nne}, for different levels of true and supposed initial uncertainty. Difficult initial uncertainty leads to highly erroneous ICP outputs that the proposed method detects if correctly parametrized.\vspace*{-0.3cm}}
\end{figure}

\section{Conclusion}\label{sec:con}
This paper presents a novel method for real time estimation of  3D uncertainty covariance matrix of the ICP algorithm. The method relies on a careful study of the influence of both sensor noises and algorithm initialization on the ICP estimates, that we leverage in a deterministic scheme which remains very simple in terms of parameter tuning. The core of our approach is   versatile as one can apply it to various choices of error metrics. However with point-to-point ICP the closed form part of the covariance is not valid, see \cite{bonnabelCovariance2016}. The approach is successfully validated on individual pairs of point clouds and over trajectories on challenging real datasets, where it obtains consistent results.
 Future work will address the benefit of the method for preventing ICP failures, particularly its coupling with learning-based methods, and for fusing odometry, ICP and GNSS in Kalman filtering and optimization-based schemes. 

\bibliographystyle{IEEEtran}
\bibliography{biblio}

\begin{thebibliography}{10}
\providecommand{\url}[1]{#1}
\csname url@samestyle\endcsname
\providecommand{\newblock}{\relax}
\providecommand{\bibinfo}[2]{#2}
\providecommand{\BIBentrySTDinterwordspacing}{\spaceskip=0pt\relax}
\providecommand{\BIBentryALTinterwordstretchfactor}{4}
\providecommand{\BIBentryALTinterwordspacing}{\spaceskip=\fontdimen2\font plus
\BIBentryALTinterwordstretchfactor\fontdimen3\font minus
  \fontdimen4\font\relax}
\providecommand{\BIBforeignlanguage}[2]{{%
\expandafter\ifx\csname l@#1\endcsname\relax
\typeout{** WARNING: IEEEtran.bst: No hyphenation pattern has been}%
\typeout{** loaded for the language `#1'. Using the pattern for}%
\typeout{** the default language instead.}%
\else
\language=\csname l@#1\endcsname
\fi
#2}}
\providecommand{\BIBdecl}{\relax}
\BIBdecl

\bibitem{censiAccurate2007}
A.~Censi, ``An {{Accurate Closed}}-form {{Estimate}} of {{ICP}}'s
  {{Covariance}},'' in \emph{{ICRA}}, 2007, pp. 3167--3172.

\bibitem{pomerleauReview2015}
F.~Pomerleau, F.~Colas, and R.~Siegwart, ``\BIBforeignlanguage{en}{A {{Review}}
  of {{Point Cloud Registration Algorithms}} for {{Mobile Robotics}}},''
  \emph{\BIBforeignlanguage{en}{Found. Trends in Robotics}}, vol.~4, no.~1, pp.
  1--104, 2015.

\bibitem{holzRegistration2015}
D.~{Holz}, A.~E. {Ichim}, F.~{Tombari} \emph{et~al.}, ``{Registration with the
  Point Cloud Library},'' \emph{IEEE RAM}, vol.~22, no.~4, pp. 110--124, 2015.

\bibitem{dubeOnline2017}
R.~Dube, A.~Gawel, H.~Sommer \emph{et~al.}, ``\BIBforeignlanguage{en}{An
  {{Online Multi}}-robot {{SLAM System}} for {{3D LiDARs}}},'' in
  \emph{\BIBforeignlanguage{en}{{IROS}}}, 2017, pp. 1004--1011.

\bibitem{genevaLIPS2018}
P.~Geneva, K.~Eckenhoff, Y.~Yang \emph{et~al.},
  ``\BIBforeignlanguage{en}{{{LIPS}}: {{LiDAR}}-{{Inertial 3D Plane SLAM}}},''
  in \emph{\BIBforeignlanguage{en}{{IROS}}}, 2018, pp. 123--130.

\bibitem{pomerleauNoise2012}
F.~Pomerleau, A.~Breitenmoser, M.~Liu \emph{et~al.},
  ``\BIBforeignlanguage{en}{Noise {{Characterization}} of {{Depth Sensors}} for
  {{Surface Inspections}}},'' in \emph{\BIBforeignlanguage{en}{{CARPI}}}, 2012,
  pp. 16--21.

\bibitem{landryCELLO3D2019}
D.~Landry, F.~Pomerleau, and P.~Gigu{\`e}re,
  ``\BIBforeignlanguage{en}{{{CELLO}}-{{3D}}: {{Estimating}} the {{Covariance}}
  of {{ICP}} in the {{Real World}}},'' in
  \emph{\BIBforeignlanguage{en}{{ICRA}}}, 2019.

\bibitem{pomerleauChallenging2012}
F.~Pomerleau, M.~Liu, F.~Colas \emph{et~al.},
  ``\BIBforeignlanguage{en}{Challenging {{Data Sets}} for {{Point Cloud
  Registration Algorithms}}},'' \emph{\BIBforeignlanguage{en}{IJRR}}, vol.~31,
  no.~14, pp. 1705--1711, 2012.

\bibitem{pomerleauComparing2013}
F.~Pomerleau, F.~Colas, R.~Siegwart \emph{et~al.},
  ``\BIBforeignlanguage{en}{Comparing {{ICP Variants}} on {{Real}}-world {{Data
  Sets}}: {{Open}}-source {{Library}} and {{Experimental Protocol}}},''
  \emph{\BIBforeignlanguage{en}{Auton. Robots}}, vol.~34, no.~3, pp. 133--148,
  2013.

\bibitem{pfisterWeighted2002}
S.~T. {Pfister}, K.~L. {Kriechbaum}, S.~I. {Roumeliotis} \emph{et~al.},
  ``{Weighted Range Sensor Matching Algorithms for Mobile Robot Displacement
  Estimation},'' in \emph{ICRA}, 2002, pp. 1667--1674.

\bibitem{barczykRealistic2017}
M.~Barczyk and S.~Bonnabel, ``\BIBforeignlanguage{en}{Towards {{Realistic
  Covariance Estimation}} of {{ICP}}-based {{Kinect V1 Scan Matching}}: {{The
  1D Case}}},'' in \emph{\BIBforeignlanguage{en}{{ACC}}}, 2017, pp. 4833--4838.

\bibitem{iversenPrediction2017}
T.~M. Iversen, A.~G. Buch, and D.~Kraft, ``\BIBforeignlanguage{en}{Prediction
  of {{ICP Pose Uncertainties Using Monte Carlo Simulation}} with {{Synthetic
  Depth Images}}},'' in \emph{\BIBforeignlanguage{en}{{IROS}}}, 2017, pp.
  4640--4647.

\bibitem{wangCharacterization2018}
Z.~Wang, Y.~Liu, Q.~Liao \emph{et~al.},
  ``\BIBforeignlanguage{en}{Characterization of a {{RS}}-{{LiDAR}} for {{3D
  Perception}}},'' in \emph{\BIBforeignlanguage{en}{{CYBER}}}, 2018.

\bibitem{laconteLidar2019}
J.~Laconte, S.-P. Desch{\^e}nes, M.~Labussi{\`e}re \emph{et~al.},
  ``\BIBforeignlanguage{en}{{{LiDAR Measurement Bias Estimation}} via {{Return
  Waveform Modelling}} in a {{Context}} of {{3D Mapping}}},'' in
  \emph{\BIBforeignlanguage{en}{{ICRA}}}, 2019.

\bibitem{deschaudIMLSSLAM2018}
J.-E. Deschaud, ``{{IMLS}}-{{SLAM}}: {{Scan}}-to-{{Model Matching Based}} on
  {{3D Data}},'' in \emph{{ICRA}}, 2018, pp. 2480--2485.

\bibitem{bengtssonRobot2003}
O.~Bengtsson and A.-J. Baerveldt, ``\BIBforeignlanguage{en}{Robot
  {{Localization Based}} on {{Scan}}-matching},''
  \emph{\BIBforeignlanguage{en}{Robotics and Auto. Sys.}}, vol.~44, no.~1, pp.
  29--40, 2003.

\bibitem{biberNormal2003}
P.~{Biber} and W.~{Strasser}, ``{The Normal Distributions Transform: a New
  Approach to Laser Scan Matching},'' in \emph{IROS}, 2003, pp. 2743--2748.

\bibitem{bonnabelCovariance2016}
S.~Bonnabel, M.~Barczyk, and F.~Goulette, ``On the {{Covariance}} of
  {{ICP}}-based {{Scan}}-matching {{Techniques}},'' in \emph{{ACC}}, 2016, pp.
  5498--5503.

\bibitem{prakhyaclosedform2015}
S.~M. Prakhya, L.~Bingbing, Y.~Rui \emph{et~al.}, ``\BIBforeignlanguage{en}{A
  {{Closed}}-form {{Estimate}} of {{3D ICP Covariance}}},'' in
  \emph{\BIBforeignlanguage{en}{MVA}}, 2015, pp. 526--529.

\bibitem{mendesICPbased2016}
E.~Mendes, P.~Koch, and S.~Lacroix, ``{{ICP}}-based {{Pose}}-graph {{SLAM}},''
  in \emph{{SSRR}}, 2016, pp. 195--200.

\bibitem{juliernew2000}
S.~Julier, J.~Uhlmann, and H.~{Durrant-Whyte}, ``A {{New Method}} for the
  {{Nonlinear Transformation}} of {{Means}} and {{Covariances}} in {{Filters}}
  and {{Estimators}},'' \emph{IEEE T-AC}, vol.~45, no.~3, pp. 477--482, 2000.

\bibitem{brossardUnscented2017}
M.~Brossard, S.~Bonnabel, and J.-P. Condomines,
  ``\BIBforeignlanguage{en}{Unscented {{Kalman}} filtering on {{Lie}}
  groups},'' in \emph{\BIBforeignlanguage{en}{{IROS}}}, 2017, pp. 2485--2491.

\bibitem{barfootAssociating2014}
T.~Barfoot and P.~Furgale, ``Associating {{Uncertainty With
  Three}}-{{Dimensional Poses}} for {{Use}} in {{Estimation Problems}},''
  \emph{IEEE T-RO}, vol.~30, no.~3, pp. 679--693, 2014.

\bibitem{barrauInvariant2018}
A.~Barrau and S.~Bonnabel, ``Invariant {{Kalman Filtering}},'' \emph{Ann. Rev.
  of Cont., Rob., and Auto. Sys.}, vol.~1, no.~1, pp. 237--257, 2018.

\bibitem{barrauInvariant2017}
------, ``The {{Invariant Extended Kalman Filter}} as a {{Stable Observer}},''
  \emph{IEEE T-AC}, vol.~62, no.~4, pp. 1797--1812, 2017.

\bibitem{bourmaudContinuousDiscrete2015}
G.~Bourmaud, R.~M{\'e}gret, M.~Arnaudon \emph{et~al.},
  ``\BIBforeignlanguage{en}{Continuous-{{Discrete Extended Kalman Filter}} on
  {{Matrix Lie Groups Using Concentrated Gaussian Distributions}}},''
  \emph{\BIBforeignlanguage{en}{J Math Imaging Vis}}, vol.~51, no.~1, pp.
  209--228, 2015.

\bibitem{long2013banana}
A.~W. Long, K.~C. Wolfe, M.~J. Mashner \emph{et~al.}, ``The banana distribution
  is gaussian: A localization study with exponential coordinates,''
  \emph{Robotics: Science and Systems VIII}, vol. 265, 2013.

\bibitem{gustafssonSome2012}
F.~{Gustafsson} and G.~{Hendeby}, ``{Some Relations Between Extended and
  Unscented Kalman Filters},'' \emph{IEEE T-SP}, vol.~60, no.~2, pp. 545--555,
  2012.

\bibitem{ovchinnikovWindowed2019}
G.~Ovchinnikov, A.~L. Pavlov, and D.~Tsetserukou, ``{Windowed Multiscan
  Optimization Using Weighted Least Squares for Improving Localization Accuracy
  of Mobile Robots},'' \emph{Autonomous Robots}, vol.~43, no.~3, pp. 727--739,
  2019.

\end{thebibliography}
\end{document}